\documentclass[10pt,twocolumn,letterpaper]{article}
\usepackage[pagenumbers]{cvpr} 

\usepackage{times}
\usepackage{epsfig}
\usepackage{graphicx}
\usepackage{amsmath}
\usepackage{amssymb}
\usepackage{blindtext}
\usepackage{bm}
\usepackage{color}
\usepackage{xcolor}
\usepackage{diagbox}
\usepackage{caption}
\usepackage{subcaption}

\usepackage[pagebackref,breaklinks,colorlinks]{hyperref}
\usepackage[capitalize]{cleveref}
\crefname{section}{Sec.}{Secs.}
\Crefname{section}{Section}{Sections}
\Crefname{table}{Table}{Tables}
\crefname{table}{Tab.}{Tabs.}

\begin{document}

\title{Interacting Attention Graph for Single Image Two-Hand Reconstruction}

\author{Mengcheng Li$^{1}$, Liang An$^{1}$, Hongwen Zhang$^1$, Lianpeng Wu$^2$, Feng Chen$^1$, Tao Yu$^1$, Yebin Liu$^{1}$\\
$^1$Tsinghua University \ \ $^2$Hisense Inc.
}

\maketitle

\begin{abstract}
Graph convolutional network (GCN) has achieved great success in single hand reconstruction task, while interacting two-hand reconstruction by GCN remains unexplored. In this paper, we present
\textbf{Int}eracting \textbf{A}ttention \textbf{G}raph \textbf{Hand} (IntagHand), the first graph convolution based network that reconstructs two interacting hands from a single RGB image. To solve occlusion and interaction challenges of two-hand reconstruction, we introduce two novel attention based modules in each upsampling step of the original GCN. The first module is the pyramid image feature attention (PIFA) module, which utilizes multiresolution features to implicitly obtain vertex-to-image alignment. The second module is the cross hand attention (CHA) module that encodes the coherence of interacting hands by building dense cross-attention between two hand vertices. As a result, our model outperforms all existing two-hand reconstruction methods by a large margin on InterHand2.6M benchmark. Moreover, ablation studies verify the effectiveness of both PIFA and CHA modules for improving the reconstruction accuracy. Results on in-the-wild images and live video streams further demonstrate the generalization ability of our network. Our code is available at \href{https://github.com/Dw1010/IntagHand}{https://github.com/Dw1010/IntagHand}. 
\end{abstract}

\section{Introduction}

Interacting two-hand reconstruction is one of the fundamental tasks towards manifold industrial applications such as virtual reality (VR), human-computer-interaction (HCI), robotics, holoportation, digital medicine, \etc Recently, monocular single hand pose and shape recovery has witnessed great success owing to deep neural networks~\cite{Zhou_2020_CVPR,3dhand_cvpr2019,Ge_2019_CVPR,2020Weakly,Zhang_2019_ICCV} and large scale datasets~\cite{2020FreiHAND,GANeratedHands_CVPR2018,Zimmermann_2017_ICCV,2020HOnnotate, joo2017panoptic, Moon_2020_ECCV_InterHand2.6M}. However, two-hand reconstruction is more challenging and remains unsolved for two reasons. First, severe mutual occlusions and appearance similarity confuse the feature extractors, making it difficult for networks to align hand poses with image features. 
Second, the interaction context between two hands is difficult to be effectively formulated during network design and training. 

\begin{figure}[ht!]
    \centering
    \includegraphics[width=0.95\linewidth]{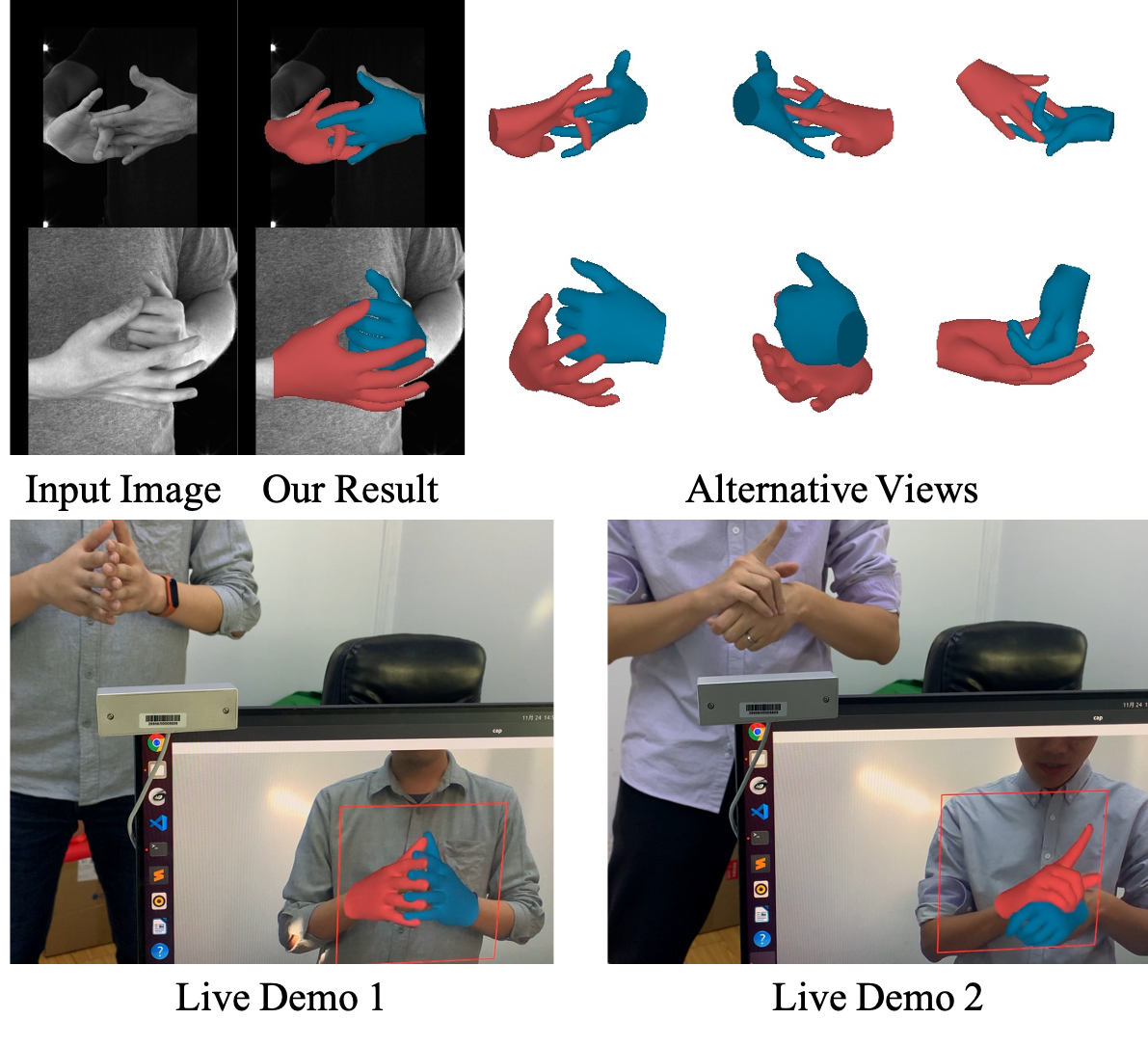}
    \vspace{-2mm}
    \caption{Illustration of our IntagHand for two-hand reconstruction. Top: results on InterHand2.6M\cite{Moon_2020_ECCV_InterHand2.6M} dataset. Bottom: real-time two-hand motion capture results on live video streams. Our method produces high quality two-hand mesh reconstruction of flexible hand poses under severe occlusions. }
    \label{fig:teaser}
    \vspace{-0pt}
\end{figure}

Monocular depth-based two-hand tracking~\cite{oikonomidis2012tracking,kyriazis2014scalable,tzionas2016capturing,taylor2017articulated,mueller2019real,wang2020rgb2hands} has been studied for years and promising results have been demonstrated. However, the energy demand and algorithm complexity restrict the ubiquitous application of depth-based methods. Recently, Wang \etal~\cite{wang2020rgb2hands} contributes a monocular RGB based two-hand reconstruction by tracking dense matching map. However, the tracking procedure itself is inherently sensitive to fast motion, and does not take full advantage of prior knowledge between interacting hands. 
Since the proposal of the large scale two-hand dataset InterHand2.6M~\cite{Moon_2020_ECCV_InterHand2.6M}, learning based single image two-hand reconstruction methods have emerged. Existing methods ~\cite{Moon_2020_ECCV_InterHand2.6M,zhang2021interacting,fan2021learning,kim2021end} either employ 2.5D heatmaps to estimate hand joint positions~\cite{Moon_2020_ECCV_InterHand2.6M,fan2021learning,kim2021end}, or use them as attention maps to extract sparse image features~\cite{zhang2021interacting}. However, such sparse local image features encoded in the heatmaps could not effectively model hand surface occlusions, and could not extract dense interaction context. 
In contrast, vertex-based
graph convolutional network (GCN) has achieved great success in single hand reconstruction~\cite{Ge_2019_CVPR,lin2021metro,lin2021mesh,tang2021towards}, yet it has not been demonstrated in two-hand conditions, and the previously mentioned challenges remain to be addressed. 

In this paper, we propose \textbf{Int}eracting \textbf{A}ttention \textbf{G}raph \textbf{Hand} (IntagHand), a novel GCN based single image two-hand reconstruction method. As a basic pipeline, we initially utilize GCN to regress mesh vertices of each hand in a coarse-to-fine manner, similar to
traditional GCN~\cite{Ge_2019_CVPR}. However, for the two-hand task, naively using a two-stream GCN to generate two hand vertices fails to utilize the interaction context between two hands, making the network confused regarding two-hand mutually occluded parts. Moreover, without any image feature feedback, the network has difficulty aligning vertices to image features as suggested by~\cite{zhang2021pymaf,lin2021mesh}. To address these issues, we equip the GCN with two novel attention modules. The first module is a pyramid image feature attention (\textbf{PIFA}) module, which uses a transformer encoder to update the latent vertex features with patched image features. Unlike projection based vertex-image alignment~\cite{zhang2021pymaf}, PIFA benefits from the global sensing ability of the attention mechanism to help each vertex seek alignment over all image patches. Furthermore, as GCN upsamples mesh vertices in a coarse-to-fine manner, we design an encoder-decoder based image feature extraction module to extract pyramid features, forcing the high resolution mesh to leverage fine-grained features. The second module is a cross hand attention (\textbf{CHA}) module that encodes interaction context into hand vertex features. The CHA module allows vertices of each hand to pay dense attention to the other hand's vertex features in order to disambiguate interhand occlusions. Benefiting from the GCN structure and the novel attention based modules, the IntagHand outperforms existing methods on InterHand2.6M~\cite{Moon_2020_ECCV_InterHand2.6M} by a large margin (\textbf{8.8}mm v.s. 13.5mm). Moreover, our method is efficient for real-time applications, producing well-aligned two-hand results on in-the-wild images and live video streams, as shown in Fig.~\ref{fig:teaser} and our project page. Overall, our contributions are summarized as: 
\begin{itemize}
    \setlength{\itemsep}{0pt}
    \item We propose the first two-hand reconstruction method using GCN based mesh regression, named IntagHand, demonstrating the effectiveness of GCN for the two-hand reconstruction task. 
    \item We propose a pyramid image feature attention (PIFA) module to distill local occlusion information with global image patch attention, producing better alignment between the hand vertices and the image features. 
    \item We propose a cross hand attention (CHA) module to implicitly model the two-hand interaction context, improving the reconstruction accuracy for closely interacting poses. 
    \item Our method achieves the new state-of-the-art results and outperforms existing solutions by a large margin on the InterHand2.6M benchmark. Furthermore, We demonstrate the generalization ability of our method on in-the-wild images. 
\end{itemize}
\section{Related Works}

\subsection{Single Hand Reconstruction}
Since the last century, hand pose estimation and gesture recognition have attracted a substantial interest~\cite{1996deformablehand,wang2013video,wang2018mask}. 
In the deep learning era, estimating the 3D hand skeleton from a single image has achieved great success~\cite{Zimmermann_2017_ICCV,GANeratedHands_CVPR2018,Cai_2018_ECCV,eccv_2020_biomechanical}. Since the proposal of the popular parametric hand model MANO~\cite{MANO:SIGGRAPHASIA:2017} and various large scale datasets~\cite{Simon_2017_CVPR,joo2017panoptic,2020FreiHAND,Moon_2020_ECCV_InterHand2.6M}, reconstructing both hand pose and shape~\cite{Zhou_2020_CVPR,Ge_2019_CVPR,baek2019pushing, 2020Weakly,Zhang_2019_ICCV,tang2021towards,lin2021metro,lin2021mesh,zhang2021hand,chen2021cameraspace} has become a mainstream approach. Among all of these methods, the most recent transformer--based models~\cite{lin2021metro,lin2021mesh} yield the best results, demonstrating the ability of the attention mechanism to learn the nonlocal relationship between any two vertices. This excellent performance inspires us to use the attention mechanism to improve mesh-image alignment and model mesh-mesh interaction. 


\subsection{Two-Hand Reconstruction}
Although nearly all single-hand reconstruction methods could extend to two-hand reconstruction tasks, few works demonstrate a result for close interacting hands. 
Two-hand reconstruction is one of the key challenges for human total motion capture. Previous body and hand simultaneous reconstruction methods~\cite{Joo_2018_CVPR,xiang2019monocular,rong2020frankmocap,zhou2021monocular,choutas2020monocular,zhang2021lightweight} all treat each hand in a separate manner and thus cannot handle close hand interaction cases such as finger knots. A recent multiview tracking based method~\cite{smith2020constraining} could reconstruct high-quality interactive hand motions, however, its hardware setup is expensive, and the algorithm is time-consuming. Monocular kinematic tracking based two-hand motion estimation methods, regardless of whether a depth sensor~\cite{oikonomidis2012tracking,kyriazis2014scalable,tzionas2016capturing,taylor2017articulated,mueller2019real} or an RGB camera~\cite{wang2020rgb2hands} is incorporated, are sensitive to fast motion and possible tracking failure. However, their dense mapping strategy, which queries correspondences between hand vertices and image pixels, inspires us to seek mesh-image alignment using dense features. In contrast, deep learning based methods such as  ~\cite{Moon_2020_ECCV_InterHand2.6M,zhang2021interacting,fan2021learning,kim2021end,rong2021monocular} directly reconstruct per-frame two-hand interaction. 
Unfortunately, all of these methods either employ 2.5D heatmaps to estimate hand joint positions~\cite{Moon_2020_ECCV_InterHand2.6M,fan2021learning}, or use them as attention maps to extract sparse image features~\cite{zhang2021interacting}, or reconstruct each hand respectively and fine-tune later~\cite{rong2020frankmocap,kim2021end}. As hands are naturally 3D surfaces, sparse local image features encoded in the heatmaps may not effectively capture hand surface occlusions and hands interaction context.  
Therefore, the mentioned methods usually fail to obtain two-hand reconstructions well aligned to images. 

\begin{figure*}[ht!]
    \centering
    \includegraphics[width=0.8\linewidth]{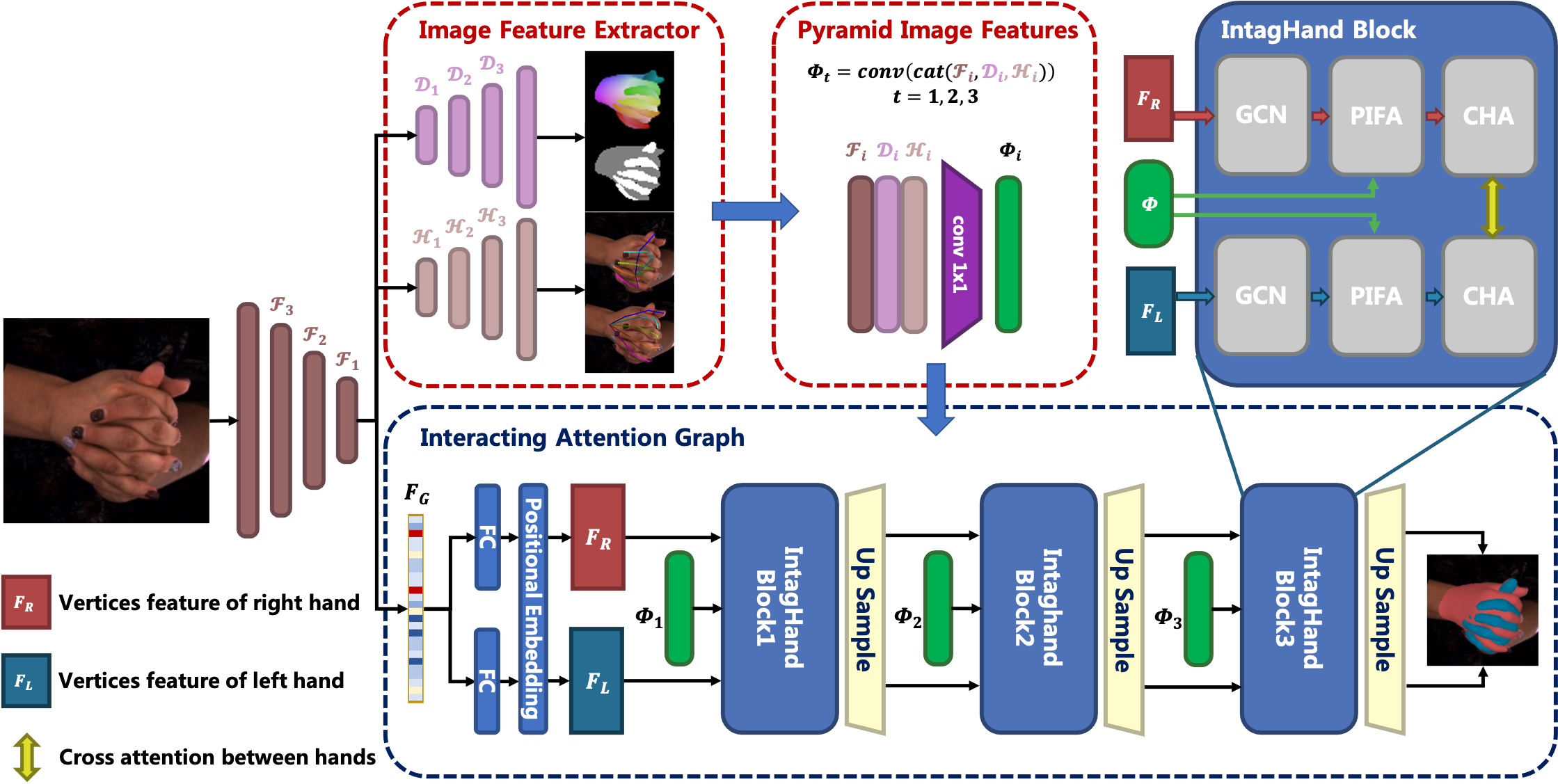}
    \caption{Our network structure. Given an RGB image as input, our network first distills a global feature vector $F_G$, a sequence of pyramid image features $\{\Phi_t, t=1,2,3\}$ along with other auxiliary predictions (2D pose, segmentation, dense mapping encoding). Then our model directly regresses the 3D coordinates of two hands surface vertices after three steps of IntagHand blocks and upsampling. Each IntagHand block contains a GCN module, a pyramid image feature attention (PIFA) module and a cross-hand attention (CHA) module. 
    }
    \label{fig:pipeline}
\end{figure*}

\subsection{Convolutional Mesh Regression}
Convolutional mesh regression (CMR), which directly regresses mesh vertices in a coarse-to-fine manner from image features using a graph convolutional network (GCN), has been proven successful for generating image-aligned 3D objects~\cite{defferrard2016gcn, wang2018pixel2mesh}, faces~\cite{ranjan2018faceautoencoder}, bodies~\cite{kolotouros2019graphcmr} or hands~\cite{Ge_2019_CVPR}. A typical CMR pipeline passes the global image feature vector through two or more cascaded graph convolution and upsampling layers and produces per-vertex 3D coordinates of the target object. Compared with joint based or rotation parameter based methods, the CMR method has denser and more semantic model representation and thus has the ability to better align image features in a per-vertex manner. However, existing CMR methods build a single forward pass without explicit image feature feedback strategy, limiting their mesh-image alignment performance as suggested by ~\cite{zhang2021pymaf}. Some recent single hand reconstruction works~\cite{lin2021mesh,tang2021towards} also employ GCN as part of the network structure; however, they discard the coarse-to-fine nature of CMR and simply use a single GCN to enhance local sensing ability.

\section{Formulation}


\subsection{Two-Hand Mesh Representation}
\label{subsec:two-hand-represent}
Unlike previous two-hand reconstruction methods~\cite{Moon_2020_ECCV_InterHand2.6M,zhang2021interacting,kim2021end,rong2021monocular} that use joints or articulated models as hand representations, we only require surface vertices with a fixed mesh topology of two hands. For convenience, we adopt the same mesh topology of the popular MANO~\cite{MANO:SIGGRAPHASIA:2017} model for each hand which contains $N=778$ vertices. To assist the attention mechanism, we define dense matching encoding for each vertex similarly to~\cite{wang2020rgb2hands} as positional embedding. Specifically, we assign different colors for different vertices while maintaining smoothness among adjacent vertices, denoted as $\{c_i\in\mathbb{R}^3, i=0,1,...,N\}$. As shown in Fig.~\ref{fig:pipeline}, our IntagHand has a hierarchical architecture that reconstructs hand mesh using three coarse-to-fine blocks, with each block followed by the upsampling layer. 
To construct the coarse-to-fine mesh topology for each block, 
we leverage the graph coarsening method introduced by \cite{dhillon2007graph_coarsen} and build $N_b=3$ level submeshes with vertex number $N_0=63, N_1=126, N_2=252$ and reserve the topological relationship between adjacent levels for upsampling. After the third block, a simple linear layer is employed to upsample the final submesh ($N_2=252$) to the full MANO mesh ($N=778$), producing the final two-hand vertices. 


\subsection{System Overview}
Our system contains two main parts: the image encoder--decoder (red dashed wireframes in Fig.~\ref{fig:pipeline}) and the interacting attention graph (blue dashed wireframe in Fig.~\ref{fig:pipeline}). 
Given a single RGB image, we first feed it to an image encoder-decoder structure that yields an intermediate global feature vector $F_G$ and several bundled feature maps $\{\Phi_t \in \mathbb{R}^{C_t \times H_t \times W_t}, t=0,1...,N_b-1\}$, where $t$ indicates that the $t^{th}$ feature level corresponds to the $t^{th}$ IntagHand block, $N_b=3$ is the block number, $H_t\times W_t$ is the resolution of the feature maps which gradually increases, and $C_t$ is the feature channel. 
Afterwards, the IntagHand takes in global feature vector $F_G$ and produces vertices of both the left and right hands. Note that each block of the IntagHand is formed by 3 submodules: a graph convolutional network (GCN) and a pyramid image feature attention module (PIFA) for each hand, together with a cross--hand attention module (CHA) between two hands. These modules are illustrated in Fig.~\ref{fig:pipeline} and will be discussed in Sec.\ref{subsec:GCN}, Sec.\ref{subsec:pymaf} and Sec.\ref{subsec:attn}, respectively. 


\begin{figure}
    \centering
    \includegraphics[width=0.75\linewidth]{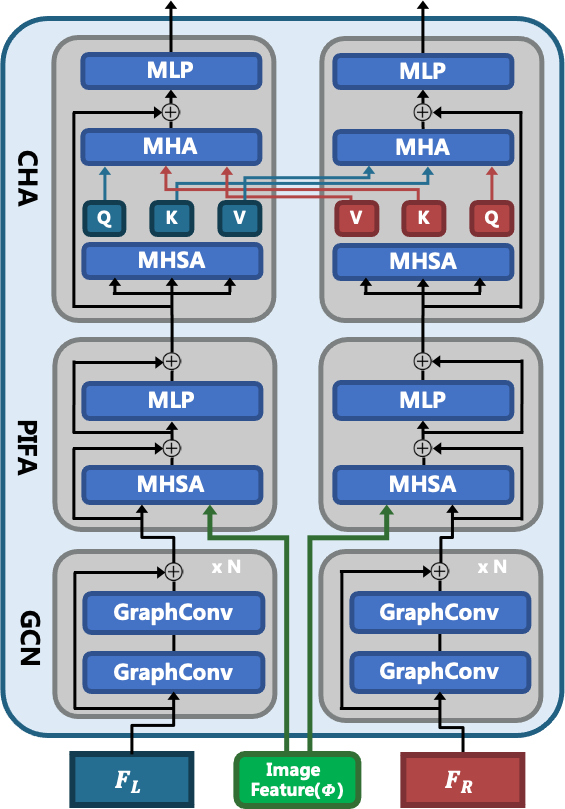}
    \caption{IntagHand Block. Our IntagHand block is formed by three parts: \textbf{1}.residual GCN module, \textbf{2}.pyramid image feature attention (PIFA) module and \textbf{3}.cross-hand attention (CHA) module. }
    \label{fig:attinblock}
    \vspace{-0pt}
\end{figure}

\section{Interacting Attention Graph}
\subsection{Graph Convolution for Two-Hand Modeling}
\label{subsec:GCN}


To directly produce two-hand vertices, our IntagHand is basically built upon previous GCN~\cite{Ge_2019_CVPR} by extending one hand stream to two-hand streams. However, different from vanilla GCN~\cite{Ge_2019_CVPR} which transforms the latent vector $F_G$ to a larger unshared per-vertex feature, we utilize fully-connected (FC) layer $g_h(\cdot)$ to map $F_G$ to a more compact feature vector $g_h(F_G)$ which is shared across vertices, and concatenate dense matching encoding (positional embedding) $c_i$ of the $i^{th}$ vertex with the shared vector to form per-vertex feature $F_V^i$ (Fig.~\ref{fig:pipeline}), which can be denoted as:
\begin{equation}
\begin{aligned}
    F_V^i = \mathrm{concat}(g_h(F_G), c_i), \\
    \quad i=0,1...,N_0; 
    \quad h=L, R,
\end{aligned}
\end{equation}
where $F_V^i \in \mathbb{R}^{f}$ is the initial graph feature, $N_0=63$ is the coarsest submesh vertex number, $f=512$ is the feature length, and $h$ indicates left (L) or right (R) hand. Such operation acts as part of the attention mechanism, and reduces the model size for faster training. 

By stacking $F_V^i$, we obtain $F_V^t \in \mathbb{R}^{N \times f}$, $t=0$. Afterwards, similar to ~\cite{Ge_2019_CVPR}, we perform the Chebyshev spectral graph CNN~\cite{defferrard2016gcn} operation (named as GraphConv for short in Fig.~\ref{fig:attinblock}) at each $t^{th} (t=0,1,2)$ block to transform input vertex features $F_V^t$ to $F_{GCN}^t$ as
\begin{equation}
    F^{t}_{GCN} = \sigma (\sum_{k=0}  ^{K-1}T_{k}^t(\hat{L^t}) F_V^t W_k^t), 
\end{equation}
where $\hat{L}^t$ is the scaled Laplacian matrix, $T_k^t$ is the $k^{th}$ term of the K-order Chebyshev polynomial, $W_k^t$ is the learnable parameter and $\sigma$ is a nonlinear activation function. $F_{GCN}^t$ denotes the intermediate features that are passed to the PIFA module. Inspired by ResNet~\cite{he2016resnet}, we add residual connection for every two GraphConv operations to assist gradient propagation and enhance learning ability; see Fig.~\ref{fig:attinblock}. 

\subsection{Pyramid Image Feature Attention Module}
\label{subsec:pymaf}

Directly reconstructing model mesh from a single global feature $F_G$ without any feedback has difficulty in guaranteeing pixel alignment with the input image~\cite{zhang2021pymaf}. Additionally, a GCN is suggested to pay more attention to local vertex features~\cite{lin2021mesh}. To solve these issues, we progressively insert hierarchical image features $\{\Phi_t \in \mathbb{R}^{C_t \times H_t \times W_t}, t=0,1,2\}$ into GCN to guarantee better mesh-image alignment using both local and global context. Note that, each image feature is a combination of both encoder feature and intermediate decoder feature for alignment to richer context (see Fig.~\ref{fig:pipeline}). Specifically, 
the output from encoder's last layer is passed through different convolutional layers to predict certain 2D information similar to~\cite{zhang2021pymaf,tang2021towards}. In our implementation, our model predicts (1) the heatmaps of joints $\mathcal{H}$, (2) the foreground mask of each hand $\mathcal{M}_L$, $\mathcal{M}_R$ and (3) the dense matching encoding of each hand $\mathcal{D}_L$, $\mathcal{D}_R$. 

To effectively use image features, we evenly divide the image feature map $\Phi_t \in \mathbb{R}^{C_t \times H_t \times W_t}$ into $N_t \times N_t$ image patches at $t^{th}$ block, and the size of each patch is $\frac{H_t}{N_t} \times \frac{W_t}{N_t}$. Then, the patches are flattened and compacted by a linear layer to yield a sequence of feature vectors $F_I^{t} \in \mathbb{R}^{(N_t \cdot N_t) \times f}$ with the same feature size $f$ to the vertex features. Afterwards, the image features $F_I^{t}$ are concatenated with vertex features $F_{GCN}^{t}$ and fed into a Multi-Head Self-Attention (MHSA) module, producing attention enhanced vertex features $F_{PIFA}^{t}$ using the following equation

\begin{equation}
\begin{aligned}
    \relax F_{PIFA}^{t} = \mathrm{MHSA}(\mathrm{concat}( F_{GCN}^t, F_I^t )). 
\end{aligned}
\label{eqn:mhsa}
\end{equation}


Although Mesh Graphformer~\cite{lin2021mesh} utilizes image feature attention (called `grid feature') as well, they use the same low--resolution image feature ($7\times7$) in the whole network while 
our image features are multi-scale ($8\times8\rightarrow16\times16\rightarrow32\times32$). 
While low--resolution image features encode more compact (or \textit{global}) information, high--resolution features contain more semantic (or \textit{local}) knowledge as they are closer to the input and output. Therefore, the pyramid structure 
forces the sparse mesh to attend to the global image features while the dense mesh to the local image features,
and could yield better vertex-image alignment.



To demonstrate the function of PIFA, we compute the attention map between the vertex domain and the image domain (please refer to ViT~\cite{dosovitskiy2020vit} for details). By adding up the PIFA attention maps of all three blocks, 
we observe that our PIFA module could distinguish between left and right hands on image pixels, and we note that PIFA pays more attention to the area of close interaction. That means PIFA module learns correct vertex-image mapping as we expect.



\begin{figure}[ht!]
    \centering
    \includegraphics[width=0.9\linewidth]{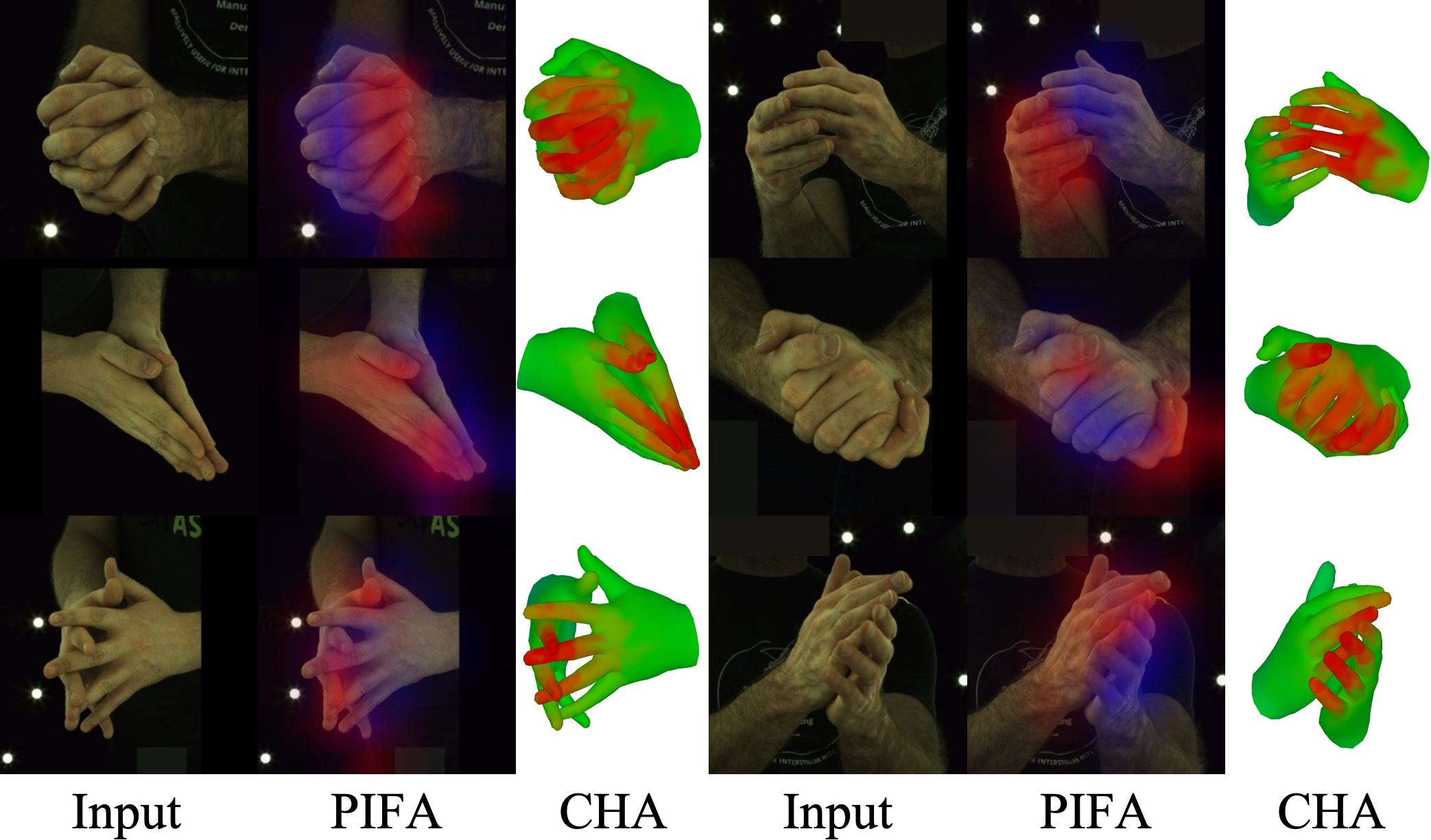}
    \caption{Visualization of attention maps in pseudo color. Six independent examples are shown. In each example, from left to right is input image, PIFA attention map overlaid on image, CHA attention map. For PIFA attention map, the {\color{red}red} represents the attention from right hand and the {\color{blue}blue} represents the left hand. The brighter color means the stronger attention.
    For CHA attention map, the redder means the stronger cross-hand attention.
    All attention maps have been normalized for better visualization.
    }
    \label{fig:attn}
    \vspace{-0pt}
\end{figure}

\subsection{Cross Hand Attention Module}
\label{subsec:attn}

It has been shown that the poses of two interacting hands are correlated~\cite{zhang2021interacting}; therefore, it is important to model hands interacting context for two-hand reconstruction. Instead of simply representing interaction as one hand's joints in the other hand's coordinate system~\cite{Moon_2020_ECCV_InterHand2.6M,zhang2021interacting}, we use a symmetric cross-hand attention (CHA) module to implicitly formulate this correlation between two hands. For simplicity, we ignore $t$ in $F_{PIFA}^t$ and use $F_L$ and $F_R$ to indicate $F_{PIFA}$ for the left and right hands, respectively.

As shown in Fig.~\ref{fig:attinblock}, we first perform MHSA on each individual hand to get $Q_h,K_h,V_h$ ($h\in{L,R}$) indicating the query, key and value feature of each hand. Then we use the query feature $Q_h$ of one hand to fetch the key feature $K_h$ and the value feature $V_h$ of the other hand through Multi-Head Attention (MHA, see Fig.~\ref{fig:attinblock}) as 




\begin{equation}
\begin{aligned}\textbf{}
   F_{R \rightarrow L}  & = \mathrm{softmax}(\frac{Q_{L} K_{R}^{T}}{\sqrt{d}}) V_R, \\
   F_{L \rightarrow R} & = \mathrm{softmax}(\frac{Q_{R} K_{L}^{T}}{\sqrt{d}})V_L,
\end{aligned}
\end{equation}
where $F_{R \rightarrow L}$ and $F_{L \rightarrow R}$ are the cross-hand attention features encoding the correlation between two hands, and $d$ is a normalization constant.
Afterwards, the cross-hand attention features are merged into the hand vertex features by a pointwise MLP layer $f_p(\cdot)$ as 

\begin{equation}
\begin{aligned}\textbf{}
    F_{L}^{'} & = f_p(F_{L} + F_{R \rightarrow L}), \\
    F_{R}^{'} & = f_p(F_{R} + F_{L \rightarrow R}),
\end{aligned}
\end{equation}
where $F_{L}^{'}$ and $F_{R}^{'}$ are the output hand vertex features, which act as $F_V^{t+1}$ of both hands for the next $t+1^{th}$ block ($t<N_b$).

It is shown in Fig.~\ref{fig:attn} that CHA also pays more attention to the closely interacting area, especially the finger-tips. This indicates that the CHA module helps to address mutual collision between hands implicitly. 

\subsection{Loss Functions}
\label{subsec:loss}

For training the image encoder-decoder, we use smooth L1 loss to supervise the 2D dense matching encoding and mean square error (MSE) loss to supervise 2D heatmaps. 


For training IntagHand, we utilize (1) vertex loss, (2) regressed joint loss and (3) mesh smooth loss. 

\noindent\textbf{Vertex Loss.} We use L1 loss to supervise the 3D coordinates of hand vertices and MSE loss to supervise the 2D projection of vertices:


\begin{equation}
\begin{aligned}
    L_{V} & = \sum_{i=1}^{N}\|V_{h,i}- V_{h,i}^{GT}\|_1 
    + \|\Pi(V_{h,i})- \Pi(V_{h,i}^{GT})\|^2_2, 
\end{aligned}
\end{equation}
where $V_{h,i}$ is $i^{th}$ vertex, $h=L,R$ means left or right hand, and $\Pi$ is the 2D projection operation, the same below. Vertex loss is applied for each submesh, which we ignore here for simplicity. 

\noindent\textbf{Regressed Joint Loss.} By multiplying the predefined joint regression matrix $\mathcal{J}$, hand joints can be regressed from the predicted hand vertices. We penalize the joint error by the following loss:
\begin{equation}
\begin{aligned}
    L_{J} & = \sum_{i=1}^{V}\|\mathcal{J} V_{h,i}- \mathcal{J} V_{h,i}^{GT}\|_1  \\
    & + \sum_{i=1}^{V}\|\Pi(\mathcal{J} V_{h,i})- \Pi(\mathcal{J} V_{h,i}^{GT})\|^2_2.
\end{aligned}
\end{equation}

\noindent\textbf{Mesh Smooth Loss.} To ensure the geometric smoothness of the predicted vertices, two different smooth losses are applied. First, we regularize the normal consistency between the predicted and the ground truth mesh:
\begin{equation}
\begin{aligned}\textbf{}
    L_{n} & = \sum_{f=1}^{F}\sum_{e=1}^{3}\|e_{f,i,h} \cdot n_{f,h}^{GT}\|_1,
\end{aligned}
\end{equation}
where $f$ is the face index of the hand mesh, $e_{f,i} (i=1,2,3)$ are the three edges of face $f$ and $n_{f}^{GT}$ is the normal vector of this face calculated from the ground truth mesh. 
Second, we minimize the L1 distance of each edge length between the predicted mesh and the ground truth mesh:
\begin{equation}
\begin{aligned}\textbf{}
    L_{e} & = \sum_{e=1}^{E}\|e_{i,h} - e_{i,h}^{GT}\|_1. 
\end{aligned}
\end{equation}

Note that, both the image encoder-decoder and the IntagHand are trained simultaneously in an end-to-end manner. 
\section{Experiments}

\begin{figure*}
    \centering
    \includegraphics[width=0.9\linewidth]{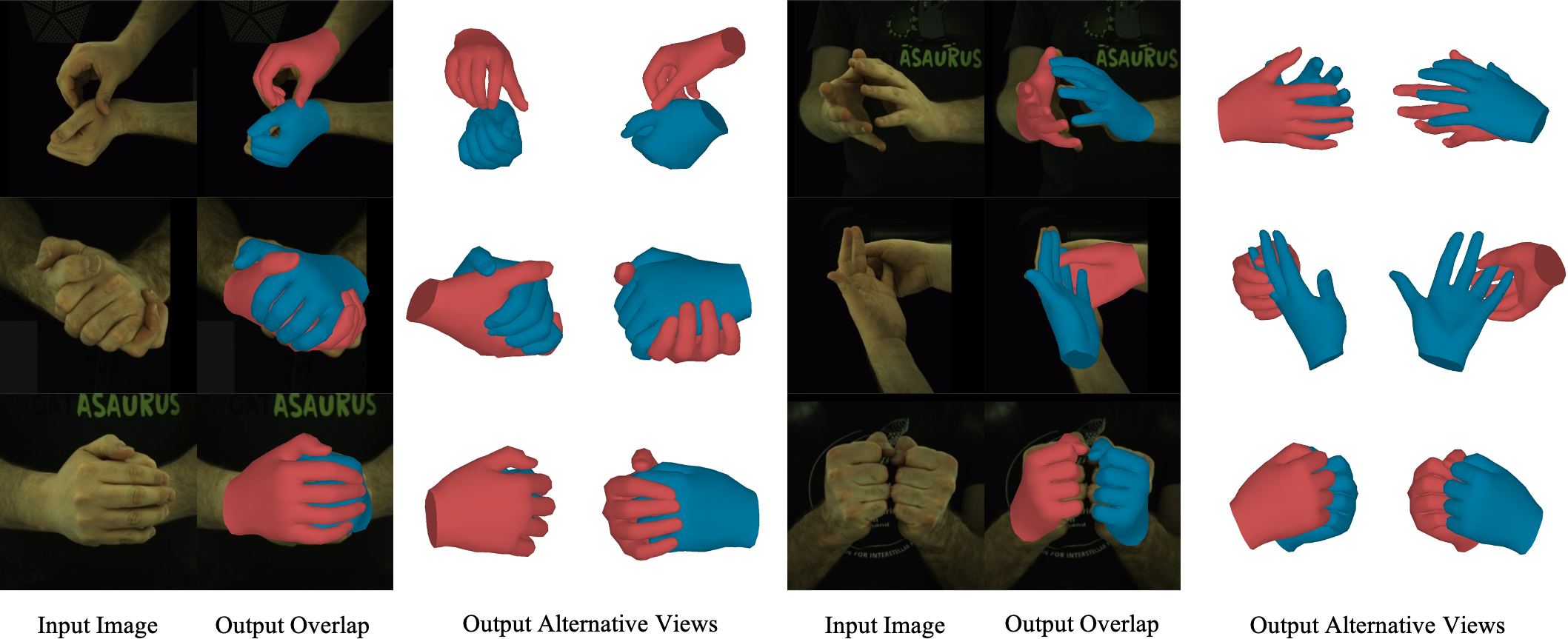}
    \caption{Qualitative results of our method on InterHand2.6M test dataset.
        Our method works well under various kinds of interactions. Note that, our method could even produce correct finger-level interactions without explicit collision detection. 
    }
    \label{fig:results}
    \vspace{-0pt}
\end{figure*}

\subsection{Experimental Settings}
\label{subsec:exp_set}

\noindent\textbf{Implementation Details.} Our network is implemented using PyTorch. We use ResNet50~\cite{he2016resnet} pretrained on ImageNet~\cite{deng2009imagenet} as backbone to encode the image feature. Following~\cite{xiao2018simple}, our image decoders utilize three simple deconvolutional layers to predict 2D joint heatmaps, 2D segmentations and dense mapping encodings. 

\noindent\textbf{Training Details.} We train our model using the Adam optimizer~\cite{kingma2014adam} on 4 NVIDIA RTX 2080Ti GPUs with the minibatch size for each GPU set as 32. The whole training takes 100 epochs across 2.5 days, with the learning rate decaying to $1\times10^{-5}$ at $50^{th}$ epoch from the initial rate $1\times10^{-4}$. During training, data augmentations including scaling, rotation, random horizontal flip and color jittering are applied. Note that, we pretrain the last upsampling layer of GCN (see Fig.~\ref{fig:pipeline}) using posed MANO meshes and fix its weights during further training. 

\noindent\textbf{Evaluation Metrics.} To evaluate both the pose and shape accuracy of reconstructed hands, we compare the Mean Per Joint Position Error (MPJPE) and Mean Per Vertex Position Error (MPVPE) in millimeters. 
For fair comparison, we follow Zhang \etal~\cite{zhang2021interacting} to scale the length of the middle metacarpal of each hand to $9.5 cm$ during training 
and rescale it back to the ground truth bone length during evaluation. This is performed after root joint alignment of each hand.
We also report the Percentage of Correct Keypoints (PCK) curve and Area Under the Curve (AUC) across linearly spanned thresholds between 0 and 50 millimeters to compare reconstruction accuracy.   

\subsection{Datasets}

\noindent\textbf{InterHand2.6M Dataset.} As the only dataset with two-hand mesh annotation, all networks in this paper are trained on InterHand2.6M \cite{Moon_2020_ECCV_InterHand2.6M} dataset\footnote{We use the v1.0\_5fps version of the InterHand2.6M which is CC-BY-NC 4.0 licensed.}.
Because we only focus on two-hand reconstruction, we pick out the interacting two-hand (IH) data with both human and machine (H+M) annotated,
and discard invalid labeling according to the $hand\_type\_valid$ annotation provided by~\cite{Moon_2020_ECCV_InterHand2.6M}. Ultimately, 366K training samples and 261K testing samples from InterHand2.6M are utilized. At preprocessing, we crop out the hand region according to the 2D projection of hand vertices and resize it to $256\times256$ resolution. 

\noindent\textbf{RGB2Hands and EgoHands Datasets.} RGB2Hands~\cite{wang2020rgb2hands} dataset consists of 4 sequences of videos with different types of two-hand interactions, and EgoHands~\cite{egohand_2015_ICCV} dataset contains 48 egocentric videos capturing complex two people interactions such as playing chess. Both datasets have no mesh annotation, therefore we only use them for qualitative evaluation. 


\subsection{Qualitative Results}
\begin{figure*}
    \centering
    \includegraphics[width=0.9\linewidth]{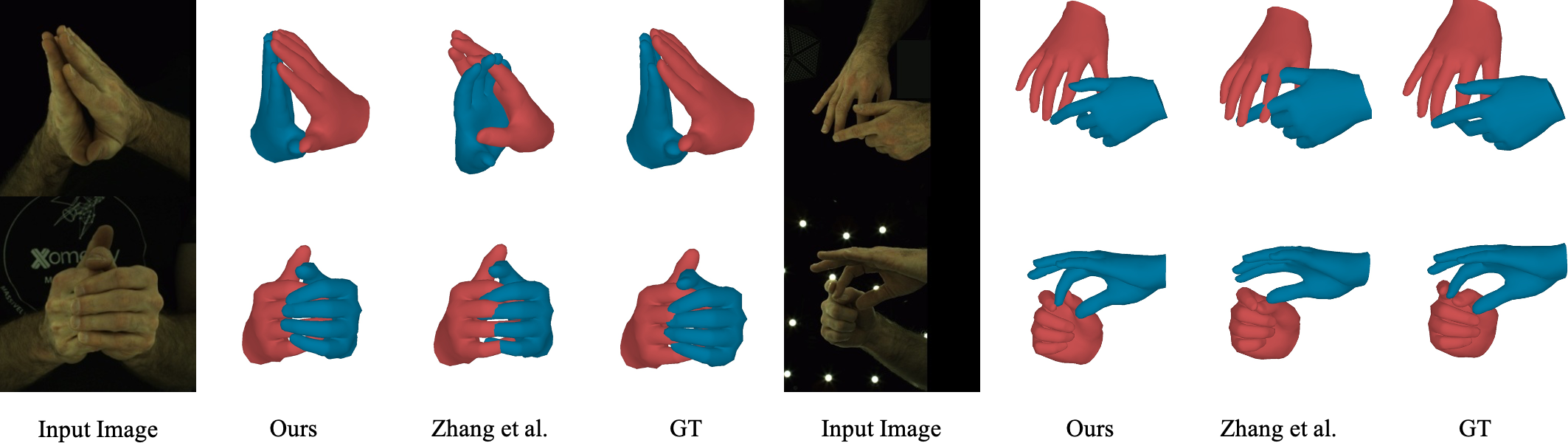}
    \caption{Qualitative comparison with Zhang \etal~\cite{zhang2021interacting} on InterHand2.6M dataset. 
    Our method produces more accurate two-hand poses, while Zhang \etal~\cite{zhang2021interacting} produces more collisions and miscalculates relative depth between the left and right hands.}
    \label{fig:compare}
    \vspace{-0pt}
\end{figure*}

Our qualitative results on InterHand2.6M~\cite{Moon_2020_ECCV_InterHand2.6M} are shown in Fig.~\ref{fig:results} and Fig.~\ref{fig:compare}. 
As shown in Fig.~\ref{fig:results}, our method could generate high quality two-hand reconstruction results under severe occlusions and various kinds of interaction context. Compared with previous state-of-the-art method~\cite{zhang2021interacting}, our method produces more realistic finger interactions and less mutual collisions of two hands (see Fig.~\ref{fig:compare}). 

\begin{figure}
    \centering
    \includegraphics[width=0.95\linewidth]{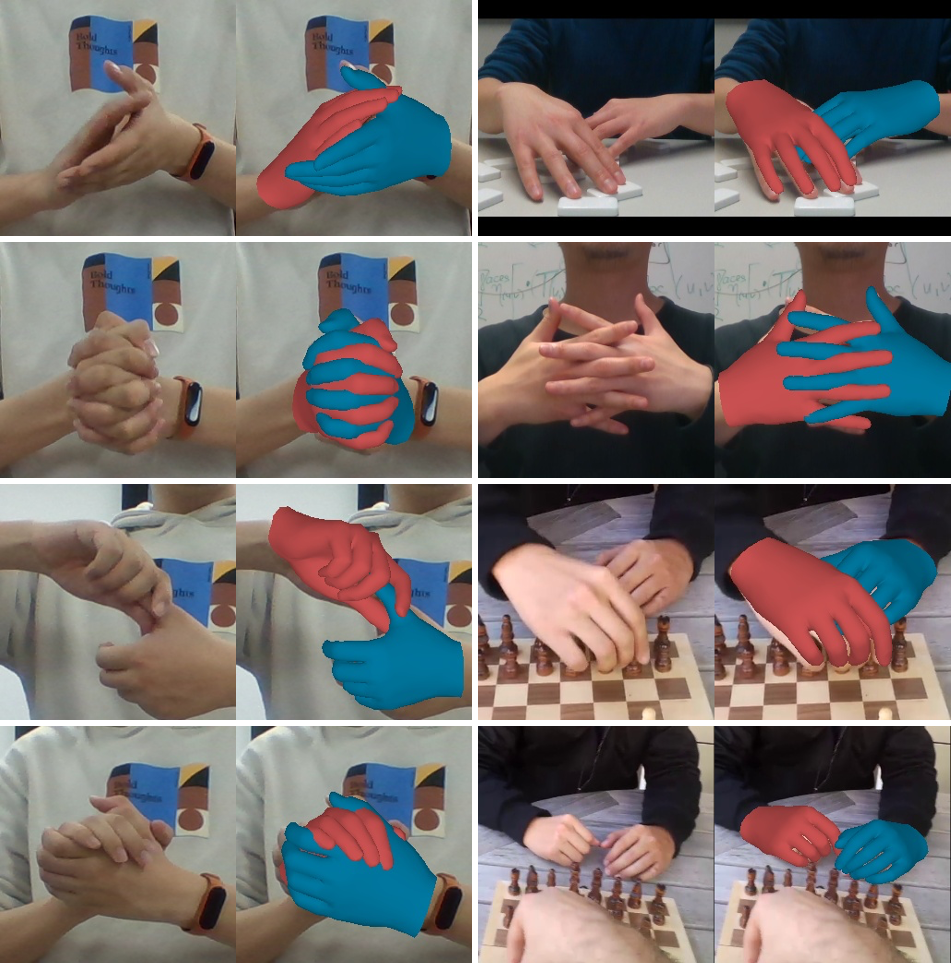}
    \caption{Qualitative results on in-the-wild images. Left 4 cases are real-life data captured using a USB camera. Top right 2 cases are taken from RGB2Hands~\cite{wang2020rgb2hands} videos. Bottom right 2 cases are taken from EgoHands~\cite{egohand_2015_ICCV} videos. 
    }
    \label{fig:real}
    \vspace{-0pt}
\end{figure}

Beyond existing methods~\cite{Moon_2020_ECCV_InterHand2.6M,zhang2021interacting,fan2021learning} which only show results in dome setting~\cite{Moon_2020_ECCV_InterHand2.6M}, we further demonstrate the generalization ability of our method on in-the-wild images. As shown in Fig.~\ref{fig:real}, our method performs well on our real-life data captured by a common USB camera. Besides, without additional training, our model yields excellent results on images from RGB2Hands dataset and EgoHands dataset, showing the potential to be applied in both third/egocentric viewpoint conditions. 
Moreover, our model runs at 30fps on single NVIDIA RTX 3090 GPU during inference, which enables future real-time applications. 


\subsection{Quantitative Comparisons}

\begin{table}
\centering
\begin{tabular}{l|cc}
\hline
  & MPJPE & MPVPE\\
\hline
$\dagger$ Zimmermann \etal \cite{Zimmermann_2017_ICCV} & 36.36 & -\\
$\dagger$ Zhou \etal \cite{Zhou_2020_CVPR} & 23.48 & 23.89\\
$\dagger$ Boukhayma \etal \cite{3dhand_cvpr2019} & 16.93 & 17.98 \\
$\dagger$ Spurr \etal \cite{spurr2018cross} & 15.40 & -  \\
\hline
Moon \etal \cite{Moon_2020_ECCV_InterHand2.6M} & 16.00 & - \\
Zhang \etal \cite{zhang2021interacting} & 13.48 & 13.95 \\
\hline 
\textbf{Ours} & \textbf{8.79} & \textbf{9.03} \\ \hline 
\end{tabular}
\caption{Comparison on InterHand2.6M. $\dagger$ means single hand methods whose results are taken from~\cite{zhang2021interacting}. We report MPJPE and MPVPE in mm, the lower the better. Our method outperforms all other methods by a huge margin.}
\label{tab:ComparisonSH}
\vspace{-0pt}
\end{table}


\begin{figure}
    \centering
    \includegraphics[width=0.80\linewidth]{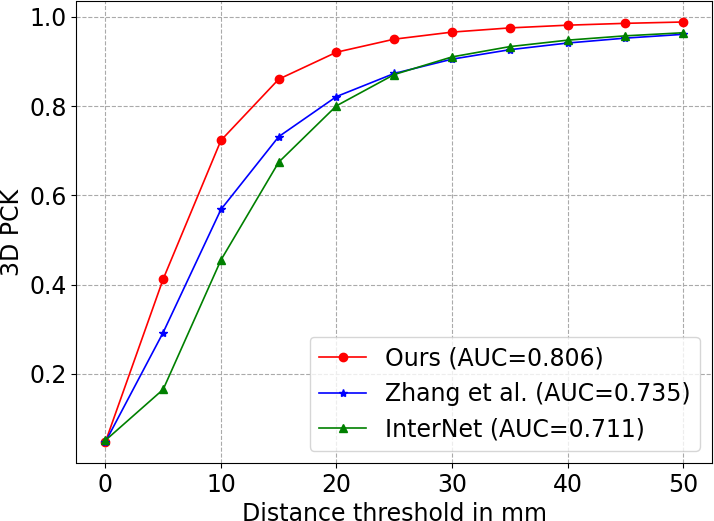}
    \caption{
    Comparison on InterHand2.6M dataset. 
    `Zhang et al' refers to ~\cite{zhang2021interacting}, while `InterNet' refers to ~\cite{Moon_2020_ECCV_InterHand2.6M}. 
    }
    \label{fig:pck}
    \vspace{-0pt}
\end{figure}

We first compare our IntagHand network with state-of-the-art single-hand reconstruction methods, as shown in Tab.~\ref{tab:ComparisonSH}. Within single-hand reconstruction methods, each hand is cropped from image by ground truth bounding box and processed separately. It is shown that reconstructing each hand individually works poorly due to heavy occlusion and appearance confusion. 

We further compare IntagHand with recent two-hand reconstruction methods. One is Moon \etal~\cite{Moon_2020_ECCV_InterHand2.6M} which regresses 3D skeletons of two hands directly. Another is Zhang \etal~\cite{zhang2021interacting}, which predicts the pose and shape parameters of two MANO~\cite{MANO:SIGGRAPHASIA:2017} models. For a fair comparison, we run their released source code on the same subset of InterHand2.6M\cite{Moon_2020_ECCV_InterHand2.6M} to that we utilize (see Sec.~\ref{subsec:exp_set}). Comparison results are shown in Tab.~\ref{tab:ComparisonSH} and Fig.~\ref{fig:pck}. It is clearly shown in Tab.~\ref{tab:ComparisonSH} that our method significantly reduces MPJPE and MPVPE. We attribute this success to the dense mesh reasoning ability of GCN and our novel attention based modules, which better align the mesh with the input image. The PCK curve in Fig.~\ref{fig:pck} further demonstrates the superior performance of our method at all error threshold levels. 


\subsection{Ablation study}

\begin{figure}
    \centering
    \includegraphics[width=0.95\linewidth]{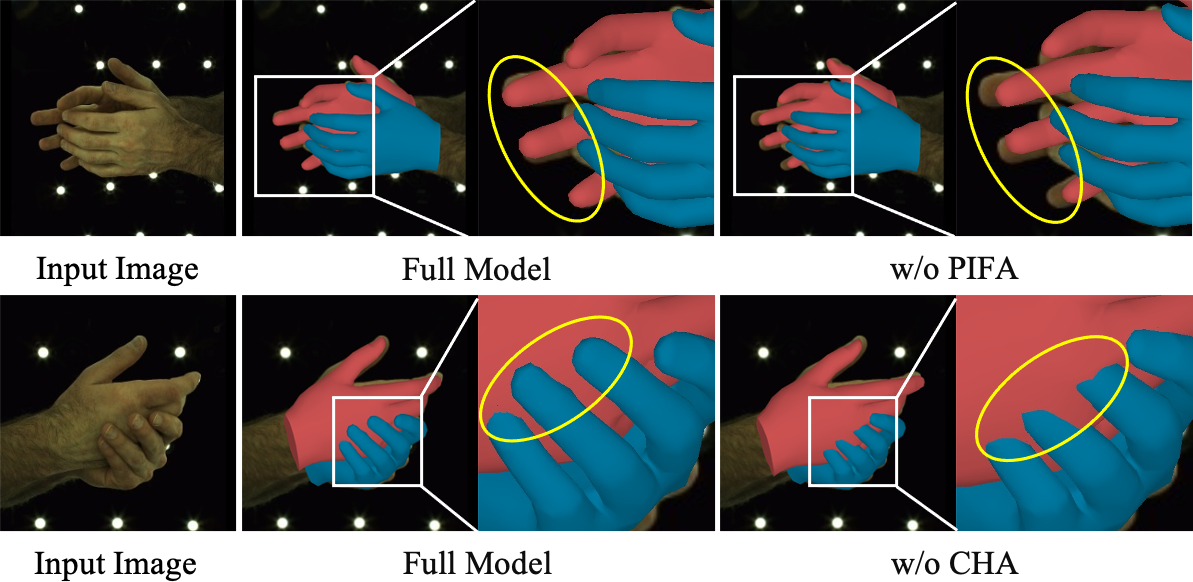}
    \caption{Qualitative ablation study on InterHand2.6M. `w/o PIFA' means removing PIFA module from the full model, `w/o CHA' means removing CHA module. It is shown that PIFA helps to align vertices and image features (top row), and CHA helps to address mutual occlusions (bottom row).}
    \label{fig:ablation}
    \vspace{-0pt}
\end{figure}

\noindent\textbf{Baseline GCN.} We train a baseline GCN model by directly modifying the GCN decoder of Ge \etal~\cite{Ge_2019_CVPR} for two-hand output (see `GCN baseline' in Tab.~\ref{tab:nopyramid}). Although directly leveraging GCN structure shows excellent numeric performance, inaccurate interaction reconstructions still exist without the attention modules. 

\noindent\textbf{Adding Attention Modules.} Based on `GCN baseline', we first add the CHA module to model interaction context (+CHA) and then add the PIFA module to further enhance vertex-mesh alignment (+CHA +PIFA), as shown in Tab.~\ref{tab:nopyramid}. By modeling interaction context with CHA, we achieve more than 0.6mm performance gain, proving the effectiveness of CHA for occlusion handling. By adding PIFA, our method further achieves more than 0.5mm performance improvement, affirming the ability of PIFA for vertex-image alignment. Qualitative comparison is shown in Fig.~\ref{fig:ablation}. 

\begin{table}
\centering
\begin{tabular}{l|cc}
\hline
                     & MPJPE  &   MPVPE    \\ 
\hline
GCN baseline  & 9.97 & 10.63\\
GCN + CHA & 9.34 & 9.59 \\
GCN + CHA + IFA-32  & 8.90 & 9.16   \\
GCN + CHA + IFA-8  & 8.83 & 9.07   \\
GCN + CHA + PIFA(\textbf{Ours})& \textbf{8.79} & \textbf{9.03} \\ 
\hline
\end{tabular}
\caption{Ablation study of module choice on InterHand2.6M.}
\label{tab:nopyramid}
\vspace{-0pt}
\end{table}

\noindent\textbf{Pyramid or Not.} 
Note that our model utilizes pyramid image features with increasing resolutions ($8\times8\rightarrow16\times16\rightarrow32\times32$). 
By removing pyramid structure, we use the consistent small ($8\times8$) or large ($32\times32$) image features in all the three IntagHand blocks (see `IFA-8' and `IFA-32' in Tab.~\ref{tab:nopyramid}). Similar to~\cite{lin2021mesh}, we find that using the small image feature performs better than using the large one. More importantly, our pyramid structure further improves the reconstruction accuracy by leveraging both the global and local information for mesh regression. 

\section{Discussion}
\noindent\textbf{Conclusion.} 
We present the interacting attention graph hand (IntagHand) method to reconstruct two interacting hands from a single RGB image. Specifically, we introduce a novel pyramid image feature attention (PIFA) module to formulate the attention relationship between hand meshes and image features, together with a novel cross-hand attention (CHA) module to encode the interaction context between two hands. Comprehensive experiments demonstrate the supreme performance of our network on InterHand2.6M dataset and in-the-wild images, and verify the effectiveness of our PIFA and CHA modules. 

\noindent\textbf{Limitation \& Impact.} The major limitation of our method is the absence of explicit mesh collision handling, resulting in occasional mesh intersections between hands. Note that, it is possible for our method to work with more than two hands where a preliminary detection network is necessary to extract hand regions and predict the hand number in each region. It is also possible to extend our method to other 2-way interactions (hand-object, human-human, \etc) as long as the subjects are encoded as vertex feature.


\noindent\textbf{Acknowledgement}: This paper is sponsored by NSFC No.62125107, NSFC No.62171255 and National Key R\&D Program of China (2021ZD0113503). 

\clearpage
{\small
\bibliographystyle{ieee_fullname}
\bibliography{ref,mvhand}
}

\end{document}